\documentclass{article}
\usepackage{spconf,amsmath,graphicx}

\usepackage{amssymb}
\usepackage{xcolor}
\usepackage{multirow}
\usepackage{graphicx} 
\usepackage{soul}
\usepackage[normalem]{ulem}
\usepackage{soul}
\sethlcolor{yellow}

\usepackage{enumitem}
\setlist{nosep, leftmargin=14pt}

\usepackage{mwe} 

\title{Understanding Annotation Error Propagation and Learning an Adaptive Policy for Expert Intervention in Barrett's Video Segmentation}

\name{Lokesha Rasanjalee$^{1,4}$ \quad Jin Lin Tan$^{2,3}$ \quad Dileepa Pitawela$^{1,4}$
\quad Rajvinder Singh$^{2,3}$ \quad Hsiang-Ting Chen$^{1,4}$}

\address{
$^{1}$ School of Computer Science and Information Technology, Adelaide University, Australia \\
$^{2}$ Adelaide Medical School, Faculty of Health and Medical Sciences, Adelaide University, Australia \\
$^{3}$ Department of Gastroenterology and Hepatology, Lyell McEwin Hospital, SA, Australia \\
$^{4}$ The Australian Institute for Machine Learning (AIML), Australia 
}

\begin{document}
%
\maketitle
\begin{abstract}
Accurate annotation of endoscopic videos is essential yet time-consuming, particularly for challenging datasets such as dysplasia in Barrett’s esophagus, where the affected regions are irregular and lack clear boundaries. Semi-automatic tools like Segment Anything Model 2 (SAM2) can ease this process by propagating annotations across frames, but small errors often accumulate and reduce accuracy, requiring expert review and correction. To address this, we systematically study how annotation errors propagate across different prompt types, namely masks, boxes, and points, and propose \textbf{Learning-to-Re-Prompt (L2RP)}, a cost-aware framework that learns when and where to seek expert input. By tuning a human-cost parameter, our method balances annotation effort and segmentation accuracy. Experiments on a private Barrett’s dysplasia dataset and the public SUN-SEG benchmark demonstrate improved temporal consistency and superior performance over baseline strategies.

\end{abstract}
\section{Introduction}
\label{sec:intro}
Developing robust AI models for endoscopic video analysis depends critically on large volumes of high-quality expert annotations. 
However, creating such datasets for less common conditions, such as dysplasia in Barrett’s esophagus, remains exceptionally challenging due to the scarcity of expert resources and the substantial time required for annotation. 
The lesions are often irregular and poorly defined, making precise annotation especially difficult.

\begin{figure}[t!]
\centering
\includegraphics[width=\columnwidth]{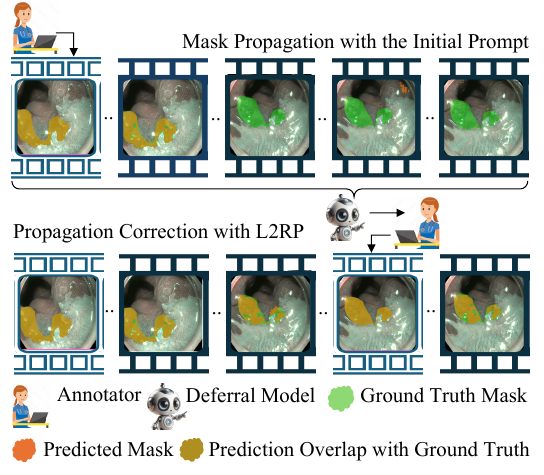}
\caption{Example Barrett’s video showing mask propagation from the initial prompt and after an L2RP-suggested correction, with improved annotation quality.
}
\label{fig:sample}
\end{figure}

Moving away from the traditional approach of annotating every frame, expert annotation can be significantly accelerated using Interactive Video Object Segmentation (iVOS) methods.
With iVOS models such as SAM2 \cite{ravi2024sam2}, experts annotate only a few key frames, while annotations for the remaining frames are automatically propagated.
These systems allow experts to interact through different prompt types, such as masks, boxes, or points that vary in precision and effort.
However, even minor segmentation errors caused by motion, lighting changes, or occlusion can accumulate over time, leading to annotation drift and requiring frequent expert correction \cite{yao2020video}.
Yet, how different prompt types affect this temporal error propagation remains unclear.
Understanding these dynamics is essential for designing systems that effectively balance segmentation accuracy against the time and effort demands placed on human experts.

In this paper, we study how segmentation errors propagate across different prompt types using a curated private Barrett’s esophagus video dataset. Building on these insights, we introduce Learning-to-Re-Prompt (L2RP), a cost-aware framework for human–AI collaboration in iVOS.
As illustrated in Fig.~\ref{fig:sample}, L2RP learns an adaptive policy that determines when to request expert input and which frame to correct, aiming to achieve the best segmentation accuracy with minimal expert intervention.
This joint formulation bridges temporal error modeling and cost-aware decision making in interactive video segmentation.
The key contributions of this work are threefold:
\begin{itemize}
     \item A systematic analysis of how segmentation errors propagate across different prompt types—masks, boxes, and points on a curated Barrett’s esophagus dataset
    \item A cost-aware framework, L2RP, that learns an adaptive policy to decide when and where expert intervention should occur.
    \item Experimental results showing that L2RP improves segmentation accuracy while reducing expert workload on a private Barrett’s dataset and a public endoscopic dataset
\end{itemize}

\section{Related Work}
\label{sec:format}
Advances in AI have achieved expert-level diagnostic accuracy when used alongside clinicians \cite{meinikheim2024influence,fockens2023deep} in Barrett’s esophagus surveillance, aiding endoscopists in detecting early neoplasia and preventing progression to esophageal cancer \cite{theocharopoulos2024deep}.
Most existing systems rely on still-image annotations, as frame-level labeling of endoscopic videos is prohibitively time-consuming.
To mitigate this, recent studies have developed scalable annotation pipelines built around iVOS models. For instance, EVA-VOS \cite{delatolas2024learning} selects high-error frames for expert correction but still overlooks the balance between segmentation gain and annotation effort.

In parallel, the Learning-to-Defer (L2D) paradigm has emerged as a principled framework for optimizing Human–AI collaboration by enabling models to selectively defer predictions to human experts when uncertainty is high \cite{mozannar2020consistent,narasimhan2022post,mao2023two,l2cu, cloc}.
However, existing L2D methods primarily address static decision-making and remain unexplored in spatiotemporal settings, such as video segmentation, where the model must determine optimal correction points that maximize segmentation accuracy while minimizing annotation effort. 

To bridge this gap, we extend L2D to a propagation-aware deferral setting and propose the L2RP framework that learns when to query correction prompts during mask propagation.
Unlike traditional L2D, where deferral denotes escalation to a more capable expert, L2RP assumes equivalent competence between prompts and focuses on identifying optimal re-prompt points to recover from propagation drift, achieving higher segmentation accuracy with minimal human intervention.

\section{Method}
\label{sec:pagestyle}

\subsection{Preliminaries}
\label{ssec:subhead}

Let \( \mathcal{V} = \{ I_{t} \}_{t=0}^{T} \) denote a Barrett’s esophagus endoscopic video sequence containing \( T{+}1 \) frames, where each frame \( I_{t} \in \mathbb{R}^{H \times W \times 3} \) represents a colour image of the mucosal surface. Every frame is annotated with a ground-truth binary mask \( M_{t} \in \{0, 1\}^{H \times W \times 1} \), where pixels labelled \( 1 \) correspond to Barrett’s dysplasia regions and \( 0 \) to normal mucosa or background.

An interactive video object segmentation model $S(\cdot)$ predicts a segmentation mask conditioned on both the video and user-provided prompts. The annotator first supplies an initial prompt $p_0$ (mask, box, or point) on the first frame $I_0$ to outline the regions of dysplasia. The model then propagates this annotation through the remaining frames, producing predicted masks $\hat{M}^{(0)} = S(\mathcal{V}, p_0)$, where each frame-level mask $\hat{M}^{(0)}_t$ represents the propagated segmentation for frame $t$. 

As the video progresses, propagated segmentation quality may deteriorate due to tissue motion, lighting changes, or endoscope movement. When such drift occurs, the expert annotator can provide a correction prompt at frame $t = \delta$, denoted $p_{\delta}$. The iVOS model would then use all prompts so far to re-propagate segmentation, $\hat{M}^{(0,\delta)} = S(\mathcal{V}, \{p_0, p_{\delta}\})$, across the sequence, where $\hat{M}^{(0,\delta)}_t$ denotes the refined mask for frame $t$.

\subsection{Learning-to-Re-Prompt}
\label{ssec:subhead}

To determine when a new correction should be requested, we introduce a \textbf{deferral model} \( D_{\theta}(\cdot) \) that learns to decide whether to continue with the current segmentation or to defer to the annotator for a new prompt. The deferral model takes as input the endoscopic video \( \mathcal{V} \) together with the propagated masks \( \hat{M}^{(0)} \) generated from the initial prompt \( p_{0} \), and outputs a discrete prediction \( d = D_{\theta}(\mathcal{V}, \hat{M}^{(0)}), \; d \in \{0, 1, 2, \ldots, T\} \). If \( d = 0 \), the model chooses \textit{not to defer} and accepts the propagated masks from the initial prompt \( p_{0} \) as final. 
Conversely, if \( d = k \) for some \( k \in \{1, 2, \ldots, T\} \), the model \textit{defers to the annotator} at frame \( k \), requesting a new correction prompt \( p_{k} \), after which the segmentation is re-propagated across the video to refine the overall mask quality.

The decision \( d \) is evaluated through a discrete, non-differentiable loss formulation inspired by the L2D framework while remaining specific to interactive video segmentation. The loss is defined as
\begin{equation}
L_{\text{def}}(d, \mathcal{V}, \hat{M}^{(0)}) = I[d = 0] \, c_{\text{prop}} + \sum_{k=1}^{T} I[d = k] \, c_{\text{corr}}^{(k)},
\label{eq:deferral_loss}
\end{equation}

where \( I[\cdot] \) denotes the indicator function. 
If \( d = 0 \), the model continues with the current segmentation and incurs the \textit{cost of accepting the initial propagation},
\(
c_{\text{prop}} = \lambda_{\text{base}} + \ell^{(0)}.
\)
If \( d = k \) for some \( k \in \{1, 2, \ldots, T\} \), the model \textit{defers to the annotator} at frame \( k \), requesting a new correction prompt \( p_{k} \), and incurs the \textit{cost of correction},
\(
c_{\text{corr}}^{(k)} = \lambda_{\text{base}} + \lambda_{\text{corr}} + \ell^{(0,k)}
\). 
Each \( \ell(\cdot) \) measures the segmentation error of the predicted masks \( \hat{M} \) for the dysplastic regions across the video, reflecting the overall propagation quality.
 The constants \( \lambda_{\text{base}} \in \mathbb{R}_{>0} \) and \( \lambda_{\text{corr}} \in \mathbb{R}_{>0} \) denote, respectively, the unavoidable cost of the initial prompt \( p_{0} \) and the additional cost of requesting a new correction during propagation. Since \( \lambda_{\text{base}} \) is common to both cases, it is omitted, while \( \lambda_{\text{corr}} \) acts as a tunable factor that balances segmentation accuracy with the additional effort required from the annotator.

Since the loss in Equation~\ref{eq:deferral_loss} is non-differentiable, we adapt the surrogate loss used in Learning-to-Defer studies~\cite{mao2023two} to our interactive video segmentation task for end-to-end training.

We keep the interactive video segmentation model \( S(\cdot) \) fixed and train only the deferral model \( D_{\theta}(\cdot) \). The deferral model takes as input the endoscopic video \( \mathcal{V} \) together with the segmentation masks \( \hat{M}^{(0)} \) propagated from the initial prompt, and outputs a series of scores \( D_{\theta}(\mathcal{V}, \hat{M}^{(0)}) = [d_{1}, d_{2}, \ldots, d_{T}] \). Each score \( d_{k} \in \mathbb{R} \) represents how suitable it is to request a correction at frame \( k \). 
To also allow the model to decide not to request a correction, we include an additional ``no-deferral'' option indexed by \( k = 0 \) with a fixed score \( d_{0} = 0 \). 

The scores are then adjusted by negating their values: we set \( \bar{d}_{0} = 0 \) for the non-deferral option and \( \bar{d}_{k} = -d_{k} \) for all \( k = 1, \ldots, T \). This means that smaller scores indicating a stronger preference to defer become larger after negation. As a result, the model treats frames that are more suitable for deferral as having higher confidence, allowing standard training losses to learn this behavior effectively. Accordingly, the surrogate loss can be defined as,
\begin{equation}
L_{\text{L2RP}}(d, \mathcal{V}, \hat{M}^{(0)}) =
\bar{c}_{\text{prop}} \, \phi(\bar{d}, 0) +
\sum_{k=1}^{T} \bar{c}_{\text{corr}}^{(k)} \, \phi(\bar{d}, k)
\label{eq:surrogate_loss}   
\end{equation}

where \( \bar{c}_{i} = 1 - c_{i} \), and \( \phi(\bar{d}, k) \) is a standard surrogate loss used for multiclass classification.
In our study, we adopt the Mean Absolute Error (MAE) as the surrogate loss $\phi$ for its stable gradients and robustness to potential class imbalance between deferral and non-deferral cases. 
The MAE-based surrogate loss can then be expressed as,
\begin{equation}
\scalebox{0.97}{$
\begin{aligned}
L_{\text{L2RP-MAE}}(d, \mathcal{V}, \hat{M}^{(0)}) 
&=\bar{c}_{\text{prop}}
\!\left[1 - \frac{1}{1 + \sum_{i=1}^{T} e^{-d_i}}\right] \\
&+ 
\sum_{k=1}^{T}
\bar{c}_{\text{corr}}^{(k)}
\!\left[1 - \frac{e^{-d_k}}{1 + \sum_{i=1}^{T} e^{-d_i}}\right]
\end{aligned}
$}
\end{equation}

During inference, the deferral model selects the action based on these scores: 
if $d_0 < \min_{1 \leq k \leq T} d_k$, the model continues without deferral; 
otherwise, it defers at frame 
$k^{*} = \arg \min_{1 \leq k \leq T} d_k$, 
where a new correction prompt $p_{k^{*}}$ is requested to refine the segmentation.

\begin{figure}[h!]
\centering
\includegraphics[width=\columnwidth]{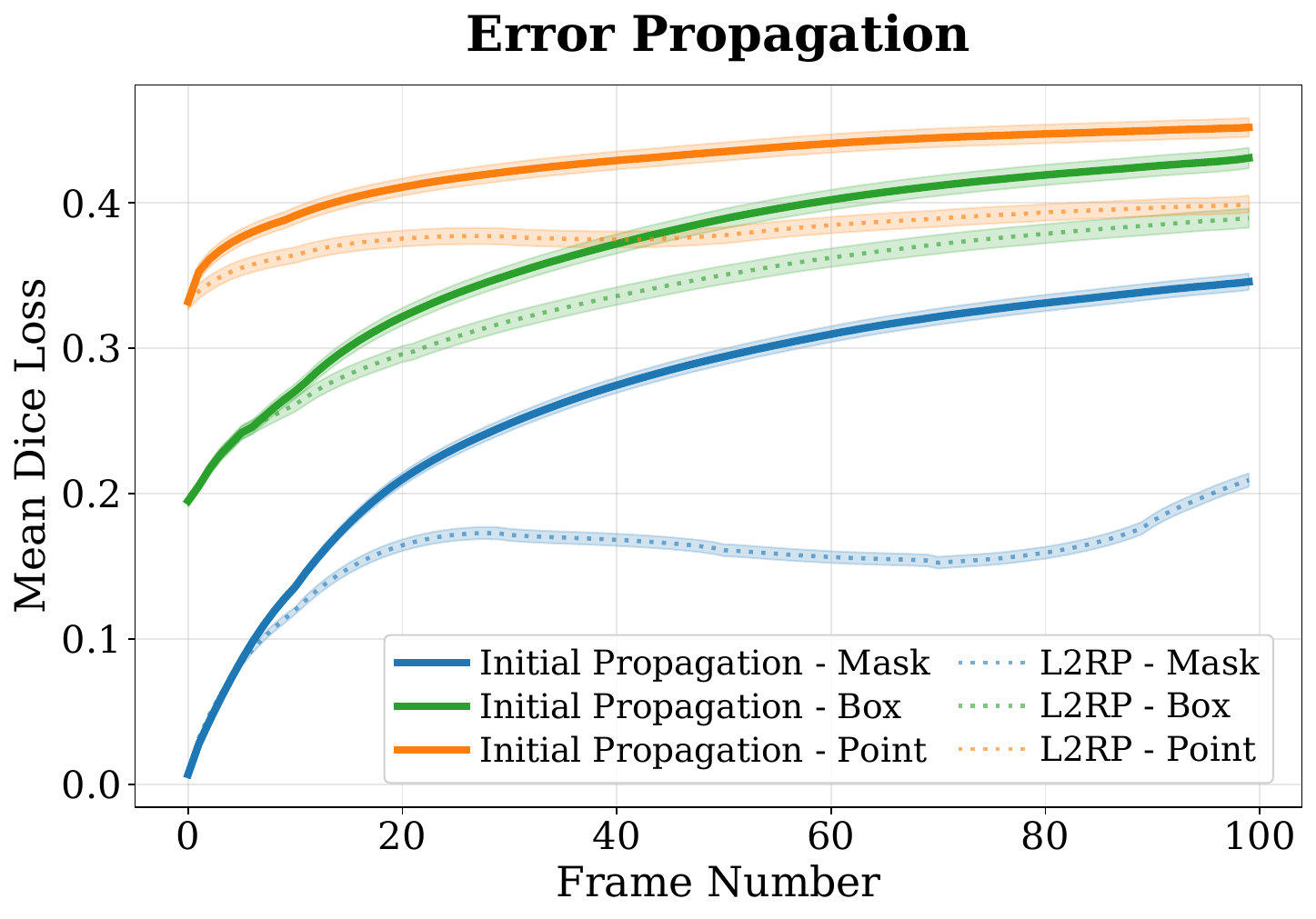}
\caption{Mean Dice Loss averaged across all videos at each frame, showing how segmentation error changes over time for different prompt types. Smoothed (20-frame window). Shaded regions indicate 95\% confidence intervals. }
\label{fig:error_prop}
\end{figure}

While \( D_{\theta} \) is originally defined over the entire video \( \mathcal{V} \) and its propagated masks \( \hat{M}^{(0)} \), in practice, we evaluate it only on an evenly spaced subset of frames $\mathcal{J}$ to avoid redundancy from similar neighboring frames. Each sample therefore includes \( \{ I_k, \hat{M}^{(0)}_k: k \in \mathcal{J} \} \), keeping the temporal context intact while reducing computation.

\begin{table*}[h!]
\centering
\resizebox{\textwidth}{!}{
\begin{tabular}{lcccccc}
\hline
\textbf{Method} & \multicolumn{3}{c}{\textbf{Barrett's}} & \multicolumn{3}{c}{\textbf{SUN-SEG}} \\
\cline{2-7}
 & Mask & Box & Point & Mask & Box & Point \\
\hline
Initial Propagation & $0.7371 \pm 0.0383$ & $0.6323 \pm 0.0645$ & $0.5477 \pm 0.0665$ & $0.5466 \pm 0.0225$ & $0.5250 \pm 0.0820$ & $0.4726 \pm 0.0595$ \\
Midpoint & $0.7988 \pm 0.0196$ & $0.6448 \pm 0.0617$ & $0.5928 \pm 0.0664$ & $0.6820 \pm 0.0190$ & $0.5363 \pm 0.0560$ & $0.5686 \pm 0.0532$ \\
Random & $0.7899 \pm 0.0241$ & $0.6459 \pm 0.0604$ & $0.5823 \pm 0.0648$ & $0.6697 \pm 0.0200$ & $0.5503 \pm 0.0560$ & $0.5421 \pm 0.0508$ \\
EVA-VOS \cite{delatolas2024learning} & $0.8244 \pm 0.0234$ & $0.6537 \pm 0.0565$ & $0.6184 \pm 0.0647$ & $0.6882 \pm 0.0244$ & $0.5701 \pm 0.0341$ & $0.5838 \pm 0.0498$ \\
\textbf{\boldmath L2RP ($\lambda_{\text{corr}} = 0.01$)} 
& $\mathbf{0.8436 \pm 0.0188^{***}}$
& $\mathbf{0.6642 \pm 0.0568^{***}}$
& $\mathbf{0.6249 \pm 0.0657^{***}}$
& $\mathbf{0.7307 \pm 0.0076^{***}}$
& $\mathbf{0.5825 \pm 0.0186^{***}}$
& $\mathbf{0.6097 \pm 0.0281^{***}}$ \\
\hline
\end{tabular}
}
\caption{
Deferral strategy comparison: Dice score (mean~$\pm$~std) across prompt types on Barrett’s and SUN-SEG datasets.
$^{***}$~Statistically significant improvement over EVA-VOS (paired Wilcoxon signed-rank test, $p < 10^{-6}$).}
\label{tab:dice_scores_barretts_sun}
\end{table*}

\section{Experiment Setting}
\label{sec:typestyle}

\textbf{Datasets:}
\label{ssec:subhead}
The in-house \textbf{Barrett’s Esophagus} dataset includes 42 videos from 16 patients, with expert pixel-level dysplasia annotations.
Consecutive 100-frame clips were extracted using a sliding window (stride${=}2$), yielding 7,599 clips. We evaluate the dataset using four-fold cross-validation with patient-level separation.
The \textbf{SUN-SEG} dataset \cite{sun_dataset,ji2022video} contains 1,106 colonoscopy videos for polyp segmentation. 
Following its original \textit{train} and \textit{test-hard} splits, we created 4,280 training and 2,500 testing clips using the same sampling method described above.

\noindent\textbf{Implementation and Training:}
We use SAM2 \cite{ravi2024sam2} as the base segmentation model $S(\cdot)$ with two memory-attention layers. 
The deferral model $D_{\theta}(\cdot)$ is a R(2+1)D network \cite{tran2018closer} pre-trained on the Kinetics-400 dataset \cite{kay2017kinetics}.
For each video, the model processes 10 frames ($|\mathcal{J}|{=}10$) with their corresponding propagated masks from the initial prompt.
Training uses the Adam optimizer (lr${=}1{\times}10^{-7}$, batch${=}16$) for 1000 epochs on an NVIDIA A100 (40 GB). We evaluate three types of interactive prompts: mask, box, and point. Masks precisely trace the dysplasia region, boxes are drawn tightly around its visible boundary, and point prompts consist of three clicks per frame. The same prompt type is used for both the initial and correction stages for consistency.

\noindent\textbf{Baselines:}
We compare the proposed L2RP framework with four baseline strategies: Initial Propagation, and three frame-selection strategies—Random, Midpoint, and an adaptation of EVA-VOS ~\cite{delatolas2024learning}.
In the Random strategy, correction frames are selected arbitrarily, whereas the Midpoint strategy always defers to the temporal center of the video. For EVA-VOS, only its frame-selection rule is used to determine when to request a correction.

\section{Results and Discussion}
\subsection{Difference in Prompt Types and Error Propagation}
Fig.~\ref{fig:error_prop} shows how segmentation errors gradually increase as annotations are propagated across video frames. When using mask prompts, the model starts with the most accurate segmentation, but the error grows quickly as the video progresses, mainly because detailed boundaries are more sensitive to small appearance changes and camera movement. Box prompts begin with slightly lower accuracy, but their errors rise more gradually, while point prompts remain the most stable over time. Interestingly, the errors of box and point prompts converge toward similar levels in later frames, indicating that their long-term propagation quality becomes comparable. Overall, detailed masks provide the best initial performance but degrade faster, whereas simpler prompts remain steadier across the sequence.

This suggests that while detailed mask annotations can produce very accurate results, they potentially require frequent corrections and therefore higher expert effort. In contrast, simpler prompts, especially points, provide more consistent results with much less manual work. When expert time is limited, using points may offer the best balance between accuracy and annotation effort.

\subsection{Efficient Expert Intervention through L2RP}
As shown in Table~\ref{tab:dice_scores_barretts_sun}, L2RP achieves the highest mean Dice scores for all prompt types, showing consistent and robust gains over baseline methods. Compared to propagation without correction, it yields substantial relative improvements of approximately +14.5\% on Barrett’s and +33.7\% on SUN-SEG for mask prompts, demonstrating its effectiveness in mitigating temporal error drift. While Midpoint and Random strategies offer only limited benefits, as they fail to adapt to segmentation errors, the EVA-VOS baseline performs competitively but remains below L2RP.

For Barrett’s dysplasia annotation workflows, these results are particularly meaningful. Given the limited budget for expert intervention, L2RP effectively optimizes the trade-off between intervention cost and propagation accuracy. By selectively identifying the most informative frames for correction, it maintains high and temporally consistent segmentation quality, ensuring that each expert correction yields a measurable improvement in performance. This adaptive mechanism leads to a more efficient balance between human effort and model accuracy in real clinical annotation settings.

\subsection{Clinical Interpretation and Practical Use of \( \lambda_{\text{corr}} \)}
In clinical data annotation workflows, the correction cost parameter $\lambda_{\text{corr}}$ represents the relative effort or time penalty of requesting additional human input to reduce error propagation.
A smaller $\lambda_{\text{corr}}$ encourages more frequent deferrals, suitable when expert intervention is inexpensive or when propagated errors escalate quickly, such as in video clips with high motion or deformation.
Conversely, a larger $\lambda_{\text{corr}}$ makes the model more conservative, prompting corrections only when predicted masks substantially deviate from lesion boundaries. This setting is preferable in high-throughput review scenarios where expert time is constrained.

Empirically, the model exhibits predictable sensitivity to this parameter. For mask prompts, Dice performance gradually decreases as $\lambda_{\text{corr}}$ increases—\textbf{0.8436 $\rightarrow$ 0.8279 $\rightarrow$ 0.8015} for $\lambda_{\text{corr}} = 0.01, 0.06, 0.08$, indicating that higher correction costs lead to fewer deferrals and consequently lower segmentation accuracy. 
In practice, $\lambda_{\text{corr}}$ can be tuned on a small validation set to balance annotation effort and accuracy, enabling users to adapt the model to their available resources.

\section{Conclusion}

In this paper, we present an analysis of annotation error propagation in interactive video segmentation and propose a novel L2RP framework that adaptively determines when to seek expert input.
Our experiments showed that while mask prompts yield the highest initial accuracy, they degrade fastest over time, whereas point prompts offer the best balance between stability and annotation effort. By dynamically selecting informative frames for correction, L2RP effectively mitigates temporal drift and optimizes the trade-off between segmentation accuracy and human effort.
Evaluation on both private and public datasets demonstrated its generalizability beyond Barrett’s videos. Overall, explicitly modeling temporal error dynamics and annotation cost enables the design of practical, annotation-efficient systems for expert-guided Barrett's dysplasia segmentation. 

\section{Compliance with Ethical Standards}
This study was performed in line with the principles of the Declaration of Helsinki. Approval was granted by the Central Adelaide Local Health Network Human Research Ethics Committee (CALHN HREC) (Date 27-04-2022 / No. 16179).

\section{Acknowledgments}
No funding was received for conducting this study. The authors have no relevant financial or non-financial interests to disclose.

\bibliographystyle{IEEEbib}
\bibliography{strings,refs}

\end{document}